\documentclass[]{article}

\usepackage[T1]{fontenc}

\usepackage[bitstream-charter]{mathdesign}
\usepackage{amsmath}
\usepackage[scaled=0.92]{PTSans}

\usepackage[
  paper  = letterpaper,
  left   = 1.25in,
  right  = 1.25in,
  top    = 1.0in,
  bottom = 1.0in,
]{geometry}
\usepackage{setspace}

\usepackage[usenames,dvipsnames]{xcolor}
\definecolor{shadecolor}{gray}{0.9}

\usepackage[english]{babel}
\usepackage[parfill]{parskip}
\usepackage{afterpage}
\usepackage{framed}
\usepackage{nicefrac}

\usepackage{graphicx}
\usepackage[labelfont=bf]{caption}
\usepackage[format=hang]{subcaption}

\usepackage{booktabs}

\usepackage{natbib}

\usepackage[algoruled]{algorithm2e}
\usepackage{algorithmic}

\usepackage[acronym,smallcaps,nowarn]{glossaries}

\usepackage[colorlinks,bookmarks=false]{hyperref}
\hypersetup{citecolor=Violet}
\hypersetup{linkcolor=black}
\hypersetup{urlcolor=MidnightBlue}

\newcommand{\be}{\begin{eqnarray}}
\newcommand{\ee}{\end{eqnarray}}

\newcommand{\n}{\nonumber \\}

\newcommand*{\qed}{\hfill\ensuremath{\blacksquare}}%

\begin{document}

\title{{\bf Sparse Probit Linear Mixed Model}}

\author{\textbf{Stephan Mandt}\footnote{Both authors have contributed equally to this work. Contact: stephan.mandt@gmail.com, wenzelfl@hu-berlin.de.}\\
  Columbia University, New York, USA\\[.2cm]
  \textbf{Florian Wenzel}$^{*}$\\
  Humboldt University of Berlin, Germany\\[.2cm]
  \textbf{Shinichi Nakajima}\\
  Technical University of Berlin, Germany\\[.2cm]
  \textbf{John Cunningham}\\
    Columbia University, New York, USA\\[.2cm]
  \textbf{Christoph Lippert}\\
  Human Longevity~Inc., Mountain View, USA\\[.2cm]
  \textbf{Marius Kloft}\\
  Humboldt University of Berlin, Germany\\
}

\maketitle 

\begin{abstract}
  Linear Mixed Models (LMMs) are important tools in statistical genetics.
  When used for feature selection, they allow
  to find a sparse set of genetic traits that best predict a continuous phenotype of interest,
  while simultaneously correcting for various confounding factors such as age, ethnicity and population structure.
  Formulated as models for linear regression, LMMs have been restricted to continuous phenotypes.
  We introduce the Sparse Probit Linear Mixed Model (Probit-LMM), 
  where we generalize the LMM modeling paradigm to binary phenotypes.
  As a technical challenge, the model no longer possesses a closed-form likelihood function.
  In this paper, we present a scalable approximate inference algorithm that
  lets us fit the model to high-dimensional data sets.
  We show on three real-world examples from different domains
  that in the setup of binary labels, our algorithm leads to
  better prediction accuracies and also selects features which show
  less correlation with the confounding factors. 
\end{abstract}

\section{Introduction}

Genetic association studies have emerged as an important branch of statistical genetics~\citep{manolio2009finding,vattikuti2014applying}. 
The goal of this field is to find causal associations between high-dimensional vectors of \emph{genotypes}, such as single nucleotide polymorphisms (SNPs), 
and observable outcomes (\emph{phenotypes, or traits}). 
For various phenotypes, such as heritable diseases, it is assumed that these associations manifest themselves on only a small number of genes.
This leads to the challenging problem of identifying few relevant positions along the genome among ten thousands of irrelevant genes.
For various complex diseases, such as bipolar disorder or type 2 diabetes \citep{craddock2010genome}, these sparse associations are largely unknown \citep{manolio2009finding},
which is why these missing associations have been entitled the \emph{The Dark Matter of Genomic Associations} \citep{dark_matter}.

Genetic associations can be spurious, unreliable, and unreproducible when the data are subject to 
spurious correlations due to confounding~\citep{imbens2015causal, pearl2009causal,morgan2014counterfactuals}.
Confounding can stem from varying experimental conditions and demographics such as age, ethnicity, or gender \citep{LiRakBor11}.
The perhaps most important types of confounding in statistical genetics arise from population structure~\citep{astle2009population}, as well as similarities between closely related samples \citep{LiRakBor11,Lip1,fusi2012joint}. Ignoring 
such confounders can often lead to spurious false positive findings that cannot be replicated on independent data \citep{kraft2009replication}.
Correcting for such confounding dependencies is considered one of the greatest challenges in statistical genetics \citep{vilhjalmsson2013nature}.

  Our approach is inspired by Linear Mixed Models (LMMs) for genome-wide association studies~\citep{Lip1}, which model the effects
  of confounding in terms of correlated noise on the traits.
  A related tool for feature selection is the LMM-Lasso~\citep{Rakitsch13}.
In this paper, we extend
the idea of LMMs to binary labels.
The LMM and its Lasso version are restricted to the
linear regression case where the output variable is continuous, but in many important applications
the phenotype is binary, such as the presence or absence of a heritable disease. To this end,
we threshold the output through a Probit likelihood~\citep{bliss1934method}. This makes parameter learning challenging
since the model becomes a Bayesian latent variable model with an intractable likelihood.
Drawing on the tools of approximate Bayesian inference, we propose two scalable inference algorithms that allow us
to fit this model to high-dimensional data.

In an experimental study on genetic data, we show that our
  approach beats several baselines. 
Compared to sparse Probit regression, our features are less correlated with the 
first principal component of the noise covariance that represents the confounder.
Furthermore, compared to the LMM-Lasso~\citep{Rakitsch13}, sparse Probit regression, and Gaussian Process (GP) classification~\citep{book:Rasmussen+Williams:2006}, our approach yields 
up to $5$ percentage points higher prediction accuracies. 
We show that our approach generalizes beyond statistical genetics in a computer malware experiment.

This paper is organized as follows. In Section~\ref{sec:2} we introduce our model and discuss related work.
Section~\ref{sec:training} then contains the mathematical details of the inference procedure. In Section~\ref{sec:experiments}
we apply our method to extract features associated with diseases and traits from confounded genetic data.
We also test our method on a data set that contains a mix of different types of 
malicious computer software data. Finally in Section~\ref{sec:conclusion} we draw our conclusions.

\section{Sparse Probit Linear Mixed Model}
\label{sec:2}
We first review the problem of confounding by population structure in statistical genetics in Section~\ref{sec:confounding}. 
In Section~\ref{sec:LMM},
we review LMMs and introduce a corresponding Probit model. We discuss the
choice of the noise kernel in Section~\ref{sec:pop} and discuss related approaches in Section~\ref{sec:connect}.

\subsection{Confounding and Similarity Kernels}
\label{sec:confounding}

The problem of confounding is fundamental in statistics.
A confounder is a common cause both of the genotypes and the traits. When it is unobserved, it induces spurious correlations
that have no causal interpretation: we say that the genotypes and traits are \emph{confounded}~\citep{imbens2015causal, pearl2009causal,morgan2014counterfactuals}. 

In statistical genetics, a major source of confounding originates from population structure~\citep{astle2009population}.
Population structure implies that due to common ancestry, individuals that are related co-inherit a large number of genes, 
making them more similar to each other, whereas individuals of unrelated ancestry obtain their genes independently, making them more dissimilar.
For this reason, collecting genetic data has to be done carefully. For example, when data are collected only in selected geographical areas (such as in specific hospitals), 
one introduces a selection bias into the sample which can induce spurious associations between phenotypes and common genes in the population.
It is an active area of research to find models that are less prone to confounding~\citep{vilhjalmsson2013nature}.
In this paper, we present such a model for the setup of binary classification.

A popular approach to correcting for confounding relies on similarity kernels, also called kinship matrices~\citep{astle2009population}.
Given $n$ samples, we can construct an $n\times n$ matrix $K$ that quantifies the similarity between samples based on some arbitrary measure. 
In the case of confounding by population structure, one
typically chooses $K_{ij} = X^\top_i X_j$, where $X_i \in {\mathbb R}^{d}$ is a vector of genetic features of individual $i$.
As $K \in {\mathbb R}^{n\times n}$ contains the scalar products between the genetic vectors of all individuals,
it is a sensible measure of genetic similarity.
As another example, when correcting for confounding by age, 
then we can choose $K$ to be a matrix that contains $1$ if two individuals have the same age, and zero otherwise. 
Details of constructing similarity kernels and other examples can be found
in~\citep{astle2009population}. 
Next, we explain how the similarity matrix can be used to correct for confounding.

\subsection{Generalizing Linear Mixed Models}
\label{sec:LMM}

We first review the LMM~\citep{henderson1950estimation},
which has been widely applied in the field of statistical genetics~\citep{fisher1919xv,yu2006unified,Lip1,Rakitsch13}.
LMMs are linear regression models that capture dependencies between the data points
in terms of correlated noise. They are a special case of generalized multivariate regression models
of the following type,
\be
\label{eq:gllm}
y_i = f \left( X_i^\top  w + \epsilon_i \,\right), \qquad \epsilon=(\epsilon_1,\ldots,\epsilon_n)^\top \sim {\cal N}(0,\Sigma),
\ee
where $f$ is an inverse link function. For LMMs, $f$ is the identity.
The outputs $y_i$ may be continuous or discrete, and $X_i$ is a set of $n$ input variables.
The variables $\epsilon_i$ are noise variables. Crucially, they are correlated and have a covariance $\Sigma$,
\be
\Sigma = \lambda_1{\bf I} +  \lambda_2 K.
\ee
The noise kernel $K$ is a modeling choice and will be discussed in Section~\ref{sec:pop}.
The noise contribution proportional to the identity matrix ${\bf I}$ is necessary to regularize the problem in case $K$ has small eigenvalues. The parameter $\lambda = (\lambda_1,\lambda_2)$ may be found by restricted maximum likelihood~\citep{patterson1971recovery}, or, as done in this work, by cross-validation. Depending on the application, we may use multiple similarity kernels. 

The crucial idea behind the model in Eq.~\ref{eq:gllm} is that parts of the observed labels
can be explained away by the correlated noise; thus not all observed phenotypes are linear effects of $X$.
By construction, the noise covariance $\Sigma$ contains information about similarities between the samples and
may be systematically used to model spurious correlations due to relatedness between samples.
The computational goal is to distinguish between these two effects.

LMMs allow to efficiently perform inference by preprocessing 
the data matrix by means of a rotation~\footnote{
  To see this, assume $f\equiv {\rm Id}$. We can always decompose the noise covariance
  as $\Sigma = U D U^\top$, where $U$ is orthogonal and $D$ is a diagonal matrix of eigenvalues of $\Sigma$.
  If we define $R = D^{-1/2}U^\top$, we can write the LMM as
  $Ry_i = R X_i^\top  w + \tilde{\epsilon_i}, \; \tilde{\epsilon}  \sim {\cal N}(0,{\bf I}).$
  Thus, after preprocessing, the remaining model is simply a linear regression model that can be treated with standard tools.
  When the inverse link function is non-linear, this methodology can not be used. In particular, we made use
  of the relation $R \circ f = f \circ R$, hence that the inverse link function commutes with the rotation.
},
which does not generalize beyond regression.
We therefore need new inference algorithms when generalizing this modeling paradigm to non-linear link functions. In this paper, we tackle inference for the important case of binary classification~\citep{bliss1934method,FahKneLanMar13}. In the following,
we  assume $f \equiv {\rm sign}$ which is the sign (or Probit) function.
This involves binary labels $y_i\in \{+1,-1\}$. 
As before, we break the independence of the label noises. This leads to the following model:
\be
\label{eq:model}
y_i = {\rm sign} \left( X_i^\top  w + \epsilon_i \,\right), \qquad \epsilon=(\epsilon_1,\ldots,\epsilon_n)^\top \sim {\cal N}(0,\Sigma).
\ee
In the special case of $\Sigma = {\bf I}$, this is just the Probit model for classification. When the noise covariance is not simply the identity but displays some non-trivial correlations, we call this modified linear mixed model the \emph{Probit Linear Mixed Model}, or short \emph{Probit-LMM}. 

Our next goal is to derive a likelihood function for our model. For the sake of a simpler notation and without loss of generality, we will assume that \emph{all observed binary labels $y_i$  are 1}.
The reason why this assumption is no constraint is that we can always perform a linear transformation 
to absorb the sign of the labels into the data matrix and noise covariance (this transformation is shown in Appendix A).
Thus, when working with this transformed data matrix and noise covariance, 
our assumption is satisfied.

Under our assumption, the likelihood function is the probability that all transformed labels are 1. 
This is satisfied when $X_i^\top w + \epsilon_i >0$. When integrating over all realizations of noise, the resulting (marginal) likelihood is

\be
{\mathbb P}(\forall i:y_i=1|w)  \; = \; {\mathbb P}(\forall i:X_i^\top w +  \epsilon_i > 0|w )\; = \;  \int_{{\mathbb R}_+^n}  {\cal N}(\epsilon; X^\top w,\Sigma) \, d^n \epsilon.
\ee
The marginal likelihood is hence an integral of the multivariate Gaussian over the positive orthant. In Section~\ref{sec:training}, we will present
efficient approximations of this integral. Before we get there, we further characterize the model.

We turn the Probit-LMM into a model for feature selection where we are interested in a point estimate of the weight vector $w$ that is sparse, i.e. 
most elements are zero. 
This is well motivated in statistical genetics, because generally only a small number of genes are believed to be causally
associated with a phenotype such as a disease.
Sparsity is achieved using the Lasso~\citep{tibshirani1996regression}, where we add an $\ell_1$-norm regularizer
to the negative marginal likelihood:
\be
 {\cal L}(w) & = & -\log \int_{{\mathbb R}_+^n}  {\cal N}(\epsilon; X^\top w,\Sigma)\,d^n\epsilon  \, + \,   \lambda_0  ||w||^1_1.
\label{eq:CorrelatedProbit}
\ee
The fact that the noise variable $\epsilon$ and the weight vector $w$ have different priors or regularizations
makes the model identifiable and lets us distinguish between linear effects and effects of correlated noise. 
In  Appendix~\ref{sec:convexity} we prove that the objective function in Eq.~\ref{eq:CorrelatedProbit} is convex.
This concludes the model; inference will be discussed in Section~\ref{sec:training}. Next, we discuss an approximation
of this model and related methods. 

\subsection{Linear Kernel and MAP Approximation}
\label{sec:pop}

We now specify the noise covariance and explore an equivalent formulation of the model.
We consider the simplest and most widely used covariance matrix $\Sigma$,
which is a combination of diagonal noise and a linear kernel of the data matrix,
\be
\Sigma & = & \lambda_1 {\bf I} + \lambda_2 X^\top X.
\ee
The linear kernel $X^\top X$ measures similarities between individuals. Since the scalar
product measures the overlap between \emph{all} genetic features, it models the dense effect of genetic similarity between samples due to population structure. 
To further motivate this kernel, we use a Gaussian integral identity: 
\be
\label{eq:global_local}
{\cal L}(w) & = & -\log \int_{{\mathbb R}_+^n}  {\cal N}(\epsilon; X^\top w, \lambda_1 {\bf I} + \lambda_2 X^\top X)\,d^n\epsilon   \; + \;   \lambda_0  ||w||^1_1\\
    & = & - \log \int_{{\mathbb R}^d} dw' \; {\cal N}(w';0,\lambda_2 {\bf I})  \int_{{\mathbb R}_+^n}  d^n \epsilon  \; {\cal N}(\epsilon; X^\top (w+w'), \lambda_1{\bf I})  \;  + \;   \lambda_0  ||w||^1_1. \n
    & = & - \log \int_{{\mathbb R}^d} dw' \; {\cal N}(w';0,\lambda_2 {\bf I})  \prod_{i=1}^n \Phi\left(\frac{X_i^\top(w+w')}{\sqrt{\lambda_1}}\right) \;  + \;   \lambda_0  ||w||^1_1. \n
    & =: &  {\cal L}_0(w) \;  + \;   \lambda_0  ||w||^1_1. \nonumber
\ee
Above, $\Phi$ is the Gaussian cumulative distribution function. We have introduced the new Gaussian noise variable $w'$.
Conditioned on $w'$, the remaining integrals factorize over $n$. However, since $w'$ is unobserved (hence marginalized out),
it correlates the samples. As such, we interpret $w'$ as a confounding variable which models the effect of the overall population on
the phenotype of interest.

The simplest approximation to the log-likelihood in Eq.~\ref{eq:global_local} is to substitute the integral over $w'$ 
by its maximum a posteriori (MAP) value:
\be
\label{eq:MAP}
{\cal L}(w,w') & = & - \sum_{i=1}^n \log \Phi\left(\frac{X_i^\top(w+w')}{\sqrt{\lambda_1}}\right) + \frac{1}{2 \lambda_2} ||w'||_2^2 + \lambda_0 ||w||_1^1.
\ee
Under the MAP approximation, the likelihood contribution to the objective function becomes completely symmetric in
 $w$ and $w'$: only the sum $w+w'$ enters. The difference between the two weight vectors $w$ and $w'$ in this approximation
is only due to the different regularizers: while $w'$ has an $\ell_2$-norm regularizer and is therefore dense, $w$
is $\ell_1$-norm regularized and therefore sparse. Every feature gets a small non-zero weight from $w'$, and only selected features get
a stronger weight from $w$. The idea is that $w'$ models the population structure, which affects all genes. 
In contrast, we are interested in learning the sparse weight vector $w$,
which has a causal interpretation because it involves only a small number of features.\footnote{Note that the interplay of two weight vectors is different from an elastic net regularizer~\citep{zou2005regularization}}.

The MAP approximation objective in Eq.~\ref{eq:MAP} is convex (proof in  Appendix~\ref{sec:convexity}) and computationally more convenient, but
is prone to overfitting. 
Under the MAP approximation we additionally optimize over $w'$, so that we can make use of the factorized form of
the objective (Eq.~\ref{eq:global_local}) over n for efficient computation.
In contrast, in the original Probit-LMM in Eq.~\ref{eq:model}, $w'$ is marginalized out. This is more expensive, but may generalize better
to unseen data. (The corresponding inference algorithm is subject of Section~\ref{sec:training}.)
We compare both approaches in Section~\ref{sec:experiments}. 

\subsection{Related Methods and Prior Work}
\label{sec:connect}

There is a large amount of literature on linear mixed models for genome-wide association studies. For a review see~\citep{price2010new,astle2009population,lippert2013linear}.
Our approach mostly relates to the 
the LMM-Lasso~\citep{Rakitsch13}.
Compared to feature selection in a simple linear regression model, the LMM-Lasso improves the selection of true non-zero
effects as well as prediction quality~\citep{Rakitsch13}.
Our model is a natural extension this model to binary outcomes, such as the disease status of a patient. While one could
also use the LMM-Lasso to model such binary labels, we show in our experimental section that this leads to
lower predictive accuracies.
As we explain in this paper, inference in our model is, however, more challenging than in \citep{Rakitsch13}.

Our model furthermore captures two limiting cases: sparse
Probit regression and GP classification~\citep{book:Rasmussen+Williams:2006}. To obtain sparse Probit regression, we simply set 
the parameters $\lambda_i = 0$ for $i \geq 2$, thereby eliminating the non-diagonal covariance structure.
To obtain GP classification, we simply omit the fixed effect (i.e., we set $w=0$) so that our model likelihood becomes
${\mathbb P}(Y=Y^{\rm obs}|w)  =  \int_{{\mathbb R}_+^n}  {\cal N}(\epsilon; 0,\Sigma) \, d^n \epsilon$, 
where the noise variable $\epsilon$ plays the role of
the latent function $f$ in GPs \citep{book:Rasmussen+Williams:2006}. When properly trained, our model 
is thus expected to outperform both approaches in terms of accuracy.
We compare our method to all three related methods in the experimental part
of the paper and show enhanced accuracy.

A common generalized linear model for classification is the logistic regression model~\citep{cox1958regression}.
Accounting for correlations in the data is non-straightforward~\citep{ragab1991multivariate};
one has to resort to approximate inference techniques, including the Laplace
and mean field approximations that have been proposed in the context of GP classification~\citep{book:Rasmussen+Williams:2006},
or the pseudo likelihood method, which has been proposed in the context of generalized LMMs~\citep{breslow1993approximate}.
To our knowledge feature selection has not been studied in a correlated logistic setup. On the other hand,
without correlations, there is a large body of work on
feature selection in Lasso regression~\citep{tibshirani1996regression}.
Alternative sparse priors to the Lasso have been suggested in~\citep{mohamed2011bayesian} for unsupervised learning
(again, without compensating for confounders).
The joint problem of sparse estimation in a correlated noise setup has been restricted to the linear regression case
\citep{seeger2011large,vattikuti2014applying,Rakitsch13}, whereas we are interested in classification.
For classification, we remark that the ccSVM~\citep{LiRakBor11} deals with confounding in a different way and it does not yield a sparse solution.
Finally, our algorithm builds on EP for GP classification~\citep{book:Rasmussen+Williams:2006,CunHenLac11}, but note that GP
classification does not yield sparse estimates and therefore does not allow us to select predictive features.

  Several alternatives to the LMM have recently been proposed 
  and shall briefly be addressed.
  \citet{song2015testing} developed a new statistical association test between traits and genetic markers. The approach reverses the placement of trait and genotype in the model and thus regresses the genotypes conditioned on the trait and an adjustment based on a fitted population structure model.
  \citet{klasen2016multi} propose a new hierarchical testing procedure, where one searches for highly correlated clusters of genotypes, and tests them for significant associations to the response variable. The significant clusters in the lowest hierarchy (or individual genotypes) are then considered as the causal genotypes of interest.
  Finally, in the context of GWAS, 
  spike-and-slap priors~\citep{carbonetto2012scalable} have been proposed as alternatives to $\ell_1$ regularizers for variable selection. In contrast to our model, where the feature weights are modeled as the sum of a dense vector $w'$ and a sparse vector $w$ contributing a small number of large effects (see Eq.~\eqref{eq:global_local}), spike-and-slap models draw each weight from exactly one of several different effect priors. While this is scalable, the approach typically results in a non-convex optimization problem. 
  Our approach has a convex optimization objective and is robust under bootstrapping, as we show in our experiments.

\section{Training Procedure}
\label{sec:training}

In this section, we lay out two efficient inference algorithms to train our model. Both algorithms
rely on approximations of the truncated Gaussian integral, which is intractable to compute in closed-form.
While the first algorithm  relies  on a point estimate for the auxiliary variable $w'$ of Eq.~\ref{eq:global_local}, the
second algorithm uses techniques from approximate Bayesian inference to estimate the truncated Gaussian integral.
While the MAP approximation algorithm is faster and easier to use in practice, the Bayesian algorithm is more precise
as we show in our experimental section.

\subsection{Prelude: ADMM algorithm}
\label{sec:prelude_ADMM} 
In both objective functions given in Eqs.~\ref{eq:global_local} and~\ref{eq:MAP}, we encounter the problem of minimizing
a convex function in the presence of an additional $\ell_1$ regularizer:
\be
{\cal L}(w) = \tilde {\cal L}(w) + \lambda ||w||_1^1. \label{eq:admm_original_objective}
\ee
(In Eq.~\ref{eq:MAP}, the objective also depends on the additional variable $w'$, in which it is smooth
and which we therefore suppress here).
The $\ell_1$-norm in the objective function is not differentiable and thus prevents us from applying standard gradient-based methods such as Newton's method.
This is a well-known problem, and several alternative solutions have been developed; one of these is the alternating direction method of multipliers (ADMM)~\citep{boyd2011distributed}.
In ADMM we augment the objective with the additional parameters $z$ and $\eta$,
\begin{align}
\begin{split}
{\cal L}(w,z,\eta)  :=  \tilde{\cal L}(w) + \lambda ||z||_1^1 + \eta^\top(w-z) +  \textstyle{\frac{1}{2}} c ||w-z||^2_2. \label{eq:add_augmented_lagrangian}
\end{split}
\end{align}
This objective can be viewed as the Lagrangian associated with the problem
\begin{align*}
  &\min_{w,z} \, \tilde {\cal L}(w)  + \lambda ||z||_1^1 +  \textstyle{\frac{1}{2}} c ||w-z||^2_2\\
  &\text{s. th. }\, z = w,
\end{align*}
which is equivalent to the original problem, Eq.~\ref{eq:admm_original_objective}. Since strong duality holds we can solve the primal problem in Eq.~\ref{eq:admm_original_objective} by solving the dual problem, Eq.~\ref{eq:add_augmented_lagrangian}. This is done by an iterative scheme where we alternate between the minimization updates for $w$ and $z$ and a gradient step in $\eta$.
Note that the term $\textstyle{\frac{1}{2}} c ||w-z||^2_2$ is optional but grants better numerical stability and faster convergence.
Details on the ADMM algorithm can be found in~\citep{boyd2011distributed}. Note that also other optimization methods are possible,
which deal with non-smooth objectives such as ours, in particular subgradient methods. The benefit of the ADMM approach,
though, is that it allows us to use second-order information because the objective is now smooth in $w$. This will be used on both
of the following algorithms.

\subsection{Maximum A Posteriori Approach} \label{sec:inference_MAP}
The simplest approximation to tackle the intractable integral relies on simply
optimizing the MAP approximated objective function of Eq.~\ref{eq:MAP}.
To this end, we minimize the objective function jointly in $(w,w')$, where
we alternate between updates in $w$ and $w'$. Cast in the form suitable for the ADMM algorithm,
the objective function becomes
\be
{\cal L}(w,w',z,\eta)  =  - \sum_{i=1}^n \log \Phi\left(\textstyle{\frac{X_i^\top(w+w')}{\sqrt{\lambda_1}}}\right) + \frac{1}{2 \lambda_2} ||w'||_2^2 + \lambda_0 ||z||_1^1 + \eta^\top(z-w).
\ee
It is straightforward to calculate the gradient in $w$ and $w'$. We do an alternating gradient descent in these variables
and carry out the additional ADMM updates in $z$ and $\eta$.

\subsection{Approximate Expectation-Maximization}

Another solution is to approximate the truncated Gaussian distribution by a simpler distribution
that allows us to solve the integral approximately. This way, we found consistent  improvements in predictive accuracy in all of our experiments. On the downside, this proposed algorithm is slightly slower in practice.

We interpret the correlated noise $\epsilon$ as a latent variable, and the sparse weights $w$ 
as global parameters. Latent variable models of this type are most conveniently solved using expectation-maximization (EM) algorithms~\citep{dempster1977maximum} that alternate between a gradient
step in the global parameters (M-step) and a Bayesian inference step (E-step) to infer the distribution over latent variables.
In our case, the E-step relies on approximate inference, which is why our approach can be called an \emph{approximate} EM algorithm.

In more detail, to follow the gradients and optimize the objective, we employ ADMM in the M-step.
Below, we derive analytic expressions for the Hessian and the gradient of the marginal likelihood
in terms of moments of the posterior distribution over the latent noise.
The inner loop (the E-step) then consists of approximating these moments by means of approximate
Bayesian inference, which we describe next. Prediction in our model is addressed in Appendix~\ref{sec:prediction}.

The inner loop of the EM algorithm amounts to computing the gradient and Hessian of ${\cal L}(w,z,\eta)$.
These are not available in closed-form, but  in terms of the first and second moment of a truncated Gaussian density.
Computing the derivatives of the linear and quadratic term is straightforward. We therefore focus on ${\cal L}_0(w)\equiv -\log \int_{{\mathbb R}_+^N}{\cal N}(\epsilon; X^\top w,\Sigma)d^n\epsilon$, which contains the intractable integral.
In the following, we use the short hand notation
\be
\mu \; \equiv \;  \mu(w) \; =\;  X^\top w.
\ee
It is convenient to introduce the following probability distribution:
\be
\label{eq:posterior}
p(\epsilon | \mu,\Sigma)  &=&  \frac{ {\mathbb 1}[\epsilon \in {\mathbb R}_+^n]\,{\cal N}(\epsilon; \mu,\Sigma) }{ \int_{{\mathbb R}_+^n}  {\cal N}(\gamma; \mu,\Sigma) \, d^n \gamma}.
\ee
Above, ${\mathbb 1}[\cdot]$ is the indicator function. This is just the multivariate Gaussian, truncated and normalized to the positive orthant. It can be considered as the Bayesian posterior of the latent multivariate noise distribution.
We furthermore introduce
\be
\label{eq:posterior-expectations}
\mu_p(w) & = &  {\mathbb E}_{p(\epsilon|\mu(w),\Sigma)} \left[\epsilon \right], \\
\Sigma_p(w) & = &   {\mathbb E}_{p(\epsilon|\mu(w),\Sigma)} \left[  (\epsilon - \mu_p(w))(\epsilon - \mu_p(w))^\top \right]. \nonumber
\ee
This is just the mean and the covariance of the \emph{truncated} multivariate Gaussian, as opposed to $\mu,\Sigma$
which are the mean and covariance of the non-truncated Gaussian. In general, these expectations do not have a closed-form solution. However, we develop suitable approximations for them in the following.

We abbreviate $\mu_p \equiv \mu_p(w)$ and $\Sigma_p\equiv \Sigma_p(w)$, and write
$\Delta \mu = \mu_p - \mu$ for the difference between the means of the posterior (the truncated Gaussian) and the un-truncated Gaussian. 
The gradient and Hessian of ${\cal L}_0(w)$ are given by 
\begin{align}
\label{eq:grad_hess}
\begin{split}
\nabla_{w}{\cal L}_0(w) & =  \Delta \mu \Sigma^{-1}  X^\top, \\
H_0(w) & =  -X [\Sigma^{-1}(\Sigma_p - \Delta \mu\Delta \mu^{\top})\Sigma^{-1} - \Sigma^{-1}]X^\top. 
\end{split}
\end{align}
Proofs are given in Appendix D. Note that the variable $w$ enters through $\Sigma_p(w)$ and $\Delta \mu(w)$.

The next step is to approximate the quantities $\mu_p$ and $\Sigma_p$ in Eq.~\ref{eq:posterior-expectations}, 
which we need for computing Eq.~\ref{eq:grad_hess}. These are intractable, involving expectations over the full posterior. Hence, we use approximate Bayesian inference methods to obtain estimates of these expectations.

A popular method for approximate Bayesian inference is Expectation Propagation (EP)~\citep{minka2001expectation}, which we use in our experimental study.
In particular, we employ EP to approximate the moments of truncated Gaussian integrals~\citep{CunHenLac11}.
EP approximates the posterior $p(\epsilon|\mu,\Sigma)$ in terms of a variational distribution $q(\epsilon)$, aiming to minimize the 
Kullback-Leibler divergence,
\be
q^*(\epsilon|\mu_{q^*},\Sigma_{q^*}) & = & {\rm arg} \min_{q} \left({\mathbb E}_p [\log p(\epsilon| \mu,\Sigma)] - {\mathbb E}_p [\log q(\epsilon|\mu_q,\Sigma_q)] \right).
\ee
The variational distribution $q^*(\epsilon)$ is an un-truncated Gaussian $q^*(\epsilon; \mu_{q^*},\Sigma_{q^*})  =  {\cal N}(\epsilon;\mu_{q^*},\Sigma_{q^*})$, 
characterized by the variational parameters $\mu_{q^*}$ and $\Sigma_{q^*}$. We approximate the posterior $p$ in terms of the variational distribution, whose mean and covariance are
$\mu_p    \approx   \mu_{q^*}$ and $\Sigma_p \approx \Sigma_{q^*}$. We
warm-start each gradient computation with the optimal parameters of
the earlier iteration.
As a remark, instead of computing the first and second moment of the
integral to compute the gradient and Hessian, the objective in Eq.~\ref{eq:CorrelatedProbit}
could also be optimized numerically using BFGS where the integral is still
approximated using EP. This is less efficient as it requires many
evaluations of the integral for a single gradient estimate.

Algorithm~\ref{alg:1} summarizes our procedure.
We denote the expectation propagation algorithm for approximating the first and second moment of the truncated Gaussian by $\text{EP}(\mu, \Sigma)$. 
Here, $\mu$ and $\Sigma$ are the mean and covariance matrix of the un-truncated Gaussian.
The subroutine returns the first and second moments of the truncated distributions $\mu_q$ and $\Sigma_q$. When initialized with the outcomes of earlier iterations,
this subroutine typically converges within a single EP loop.

\begin{algorithm}
\caption{Approximate Inference for the Probit-LMM}
\label{alg:1}
\begin{algorithmic}
\STATE pre-process the data, absorb binary labels into $X$, compute $\Sigma$.
\REPEAT

  \STATE initialize w = $w^{k}$
    \REPEAT
              \STATE $(\mu_q, \Sigma_q) \leftarrow \text{EP}(X^\top w, \Sigma)$
              \STATE $\Delta \mu = \mu_q - X^\top w$
              \STATE $g = \Delta \mu^\top \Sigma^{-1}  X^\top + c (w-z^k+\eta^k)^\top$
              \STATE $H = X [\Sigma^{-1} - \Sigma^{-1}(\Sigma_q - \Delta \mu\Delta \mu^{\top})\Sigma^{-1}]X^\top + c \mathbf{I}$
              \STATE $w = w - \alpha_t H^{-1} g$ 
      \UNTIL{criterion 2 is met}
  \STATE \emph{\textbackslash\textbackslash ADMM updates}
  \STATE $w^{k+1} = w$
  \STATE $z^{k+1} = S_{\lambda/c} (w^{k+1} + \eta^k)$   \textbackslash\textbackslash \emph{soft thresholding, see \citet{boyd2011distributed}}
  \STATE $\eta^{k+1} = \eta^k + w^{k+1} - z^{k+1}$
\UNTIL{criterion 1 is met}    
\end{algorithmic}
\end{algorithm}

Our algorithm thus consists of two nested loops; the outer ADMM loop, containing the Newton update, and the inner EP loop, which computes the moments of the posterior. We choose stopping \emph{criterion 1} to be the convergence criterion proposed by Boyd \citep{boyd2011distributed} and choose \emph{criterion 2} to be always fulfilled, i.\,e. we perform only one Newton optimization step in the inner loop. Our experiments showed that doing only one Newton optimization step, instead of executing until convergence, is stable and leads to significant improvements in speed. ADMM is known to converge even when the minimizations in the ADMM scheme are not carried out exactly (see e.g. \citep{Eckstein:1992:DSM:153390.153393}).

\section{Empirical Analysis and Applications}\label{sec:experiments}
We study the performance of our proposed methods in experiments on both artificial and real-world data.
We consider the two versions of our model: Probit-LMM (which  minimizes Eq.~\ref{eq:global_local} with respect to $w$) and Probit-LMM MAP (that minimizes Eq.~\ref{eq:MAP} with respect to both $w$ and $w'$).
Our data are taken from the domains of statistical genetics and
computer malware prediction.

We compare our algorithms against three competing methods, including sparse Probit regression, GP classification and the LMM-Lasso. In all considered cases, the Probit-LMM achieves higher classification performance. Also, the features that our algorithms find are less affected by spurious correlations induced by population structure. We find that the Probit-LMM outperforms its MAP approximation across all considered datasets. Yet, in many cases the MAP approximation is a cheap alternative to the full model.

\subsection{General Experimental Setup}\label{sec:exp_setup}
For the real-world and synthetic experiments, we first need to make a choice for the class of kernels that we use for the covariance matrix. We choose a combination of three contributions,
\be\label{eq:kernels}
\Sigma = \lambda_1 {\bf I} + \lambda_2 X^\top X + \lambda_3 \Sigma_{\text{side}}.
\ee
The third term is optional and depends on the context; it is a kernel representing any side information provided in an auxiliary feature matrix $X'$.
Here, we compute $\Sigma_{\text{side}}$ as an RBF kernel\footnote{\label{foot:rbf} The radial basis function (RBF) kernel function is defined as $k(x_1,x_2):= \exp\left( -\frac{1}{2}\sigma^{-2}||x_1-x_2||^2 \right)$, where $\sigma$ is the length scale parameter. The entries of the kernel matrix are $\left(\Sigma_{\text{side}}\right)_{ij} = k(X'_i, X'_j)$ with $X'_i, X'_j$ are the side information corresponding to data point $i$ and $j$, respectively.}
from the side information $X'$. 
Note that this way, the data matrix enters the model both through the linear effect but also through the linear kernel. 
 We evaluate the methods by using $n$ individuals from the dataset for training, and splitting the remaining dataset equally into validation and test sets. This process is repeated 50 times, over which we report on average accuracies or areas under the ROC curve (AUCs), as well as standard errors \citep{fawcett2006introduction}. 

The hyperparameters of the kernels, together with the regularization parameter $\lambda_0$, were determined on the validation set, using grid search over a sufficiently large parameter space (optimal values are attained inside the grid; in most cases $\lambda_i \in [0.1, 1000]$). For all datasets, the features were centered and scaled to unit standard deviation, except in experiment \ref{sec:malware}, where the features are binary.

In Sections~\ref{sec:tub} and \ref{sec:malware}, we show that including a linear kernel into the covariance matrix leads to top-ranked features which are less correlated with the population structure in comparison to the top-ranked features of sparse Probit regression. The correlation plots\footnote{The correlation plots in Fig.~\ref{fig:correlation} are created according to \citep{LiRakBor11} as follows. First, we randomly choose $70\%$ of the available data as training set and obtain a weight vector $w$ by training. We compute the empirical Pearson correlation coefficient
of each feature with the first principle component of the linear
kernel on top of the data. This is a way to measure the correlation
with the population structure \citep{price2006eigenstrat}. We define
the index set $I$ by taking the absolute value of each entry of $w$
and sorting them in descending order. We now sort the so-obtained list
of correlation coefficients with respect to the index set $I$ and
obtain a resorted list of correlation coefficients $(c_1, \dots ,
c_n)$. In the last step, we obtain a new list $(\hat c_1,\dots,\hat
c_n)$ by smoothing the values, computing $\hat c_i :=
\frac{1}{i}\sum_k^i c_k$. Finally, we plot the values $(\hat
c_1,\dots,\hat c_n)$ with respect to $I$. This procedure was repeated
30 times for different random choices of training sets.} in
Fig.~\ref{fig:correlation} show the mean correlation of the top
features with population structure and the corresponding
  standard errors. All experiments were performed on a linux machine with 48 CPU kernels (each 2.4GHz) and 368GB RAM.
\subsection{Simulated Data}\label{sec:toy_exp}
To test the properties of our model in a controlled setup, we first generated synthetic data as follows.
We generate a weight vector $w \in \mathbb{R}^{d}$ with $1 \le k\leq d$ entries being $1$, and the other $d-k$ entries being $0$. We chose $d=50$ and varied $k$.
We then create a random covariance matrix $\Sigma_{\text{side}} \in \mathbb{R}^{n\times n}$,
which serves as side information matrix\footnote{
  The covariance matrix was created as follows.
  The random generator in MATLAB version 8.3.0.532 was initialized to seed = 20 using the
  \texttt{\small rng(20)} command. The matrix $\Sigma_{\text{side}}$ was realized in two steps via 
  \texttt{\small A=2*rand(50,200)-1} and \texttt{\small$\Sigma_{\text{side}}$=3*A'*A+0.6*eye(200)+3*ones(200,200)}.
}.
We chose $n=200$ and drew $200$ points $X = \{x_1,\dots x_{n}\}$ independently from a uniform distribution over the unit cube $[-1,1]^{d}$ and create the labels according to the Probit model, Eq.~\ref{eq:model}, using $\Sigma_{\text{side}}$ as covariance matrix. We reserve $100$ samples for training and $50$ for validation and testing, respectively.

\begin{figure}[!h]
\begin{center}
\includegraphics[width=0.80\linewidth]{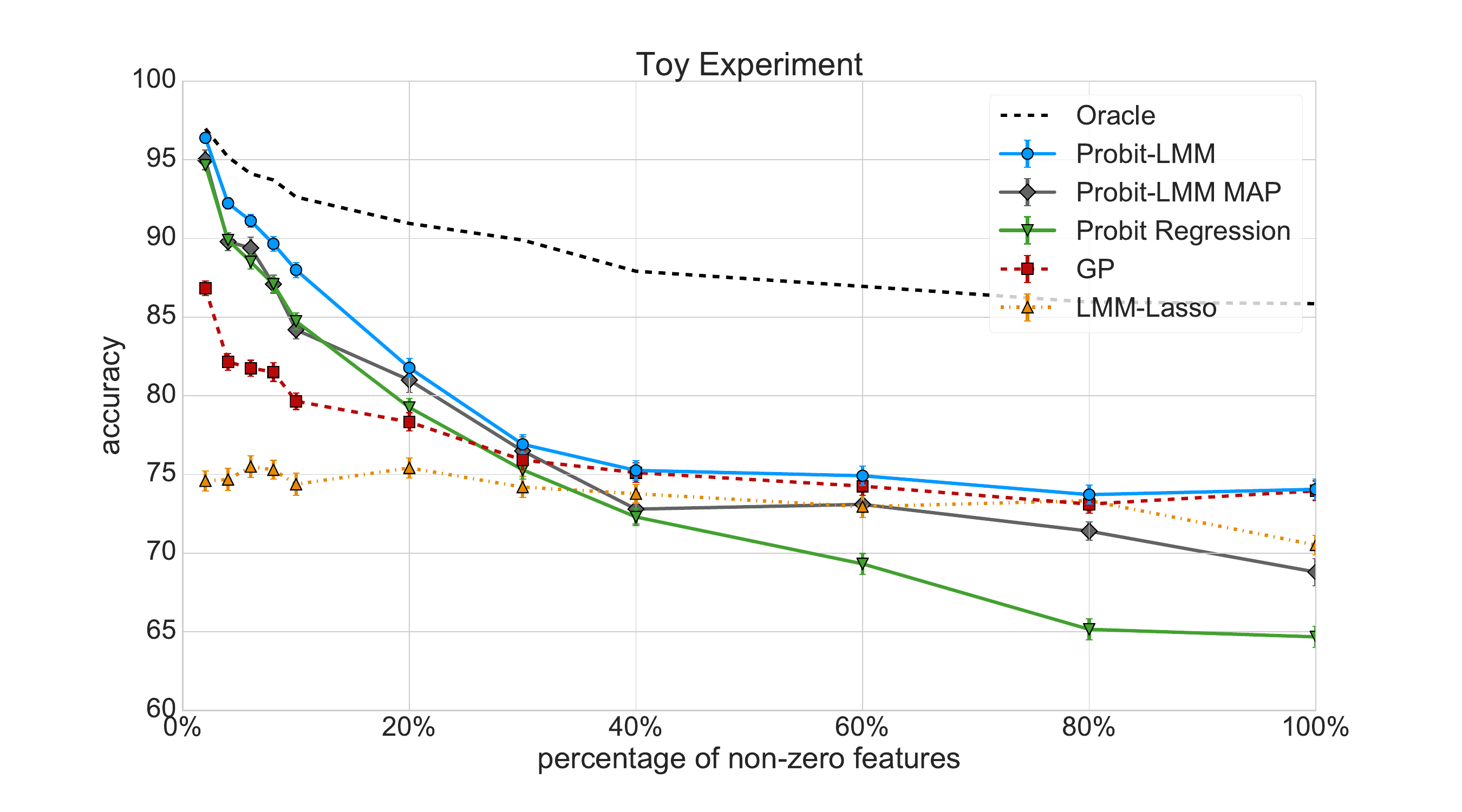} 
\caption{\textsc{Toy}: Average accuracies as a function of the number of true non-zero features in the generating model. (Proposed methods: Probit-LMM and MAP approximation)
}
 \label{fig:toy_accuracy}
\end{center}
\end{figure}

  The synthetic data allowed us to control the sparsity level $k$ of non-zero features. We then fit various models to the data to predict the
binary labels: Probit-LMM (proposed) as well as Probit-LMM MAP (proposed), GP-classification, the LMM-Lasso, and standard $\ell_1$-norm regularized (sparse) Probit
regression. As a benchmark we introduce the {\it oracle classifier}, where we use the Probit-LMM (with covariance matrix $\Sigma_{\text{side}}$), but skip the training and instead use the true underlying $w$ for prediction.
Fig.~\ref{fig:toy_accuracy} shows the resulting accuracies. The horizontal axis shows the varying percentage of non-zero features in the artificial data $k/d$.
Note that the accuracies of all methods fluctuate due to the finite size of the different data sets that we generated.

\begin{figure}[!h]
\begin{center}
\includegraphics[width=0.99\linewidth]{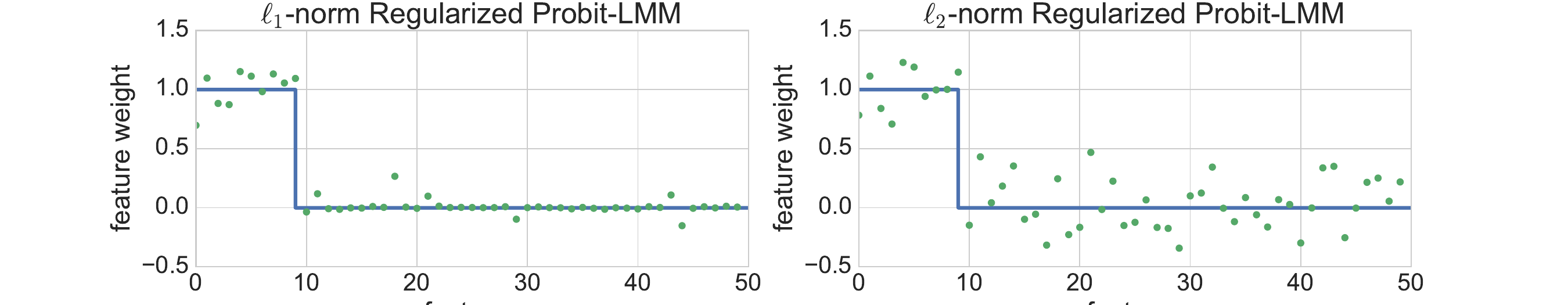} 
\caption{\textsc{Toy}: Effects of the regularizer on the model's ability to select features. Ground truth (blue solid line) and feature weights (green dots) of $\ell_1$-norm (\textsc{Left}) and $\ell_2$-norm (\textsc{Right})
         regularized Probit-LMM.  
}
 \label{fig:toy_sparsity}
\end{center}
\end{figure}

The observed performances of the methods depend on the varying level of sparsity of the data: if the true linear effect is sparse, sparsely regularized models  should be expected to work better. The opposite can be expected from
models that include all features in a dense way, such as GP classification. These models are good
when the true effects are dense. Our plot indeed reveals this tendency. 
$\ell_1$-norm regularized (sparse) Probit regression performs well for small $k$, whereas GP classification works well for large $k$.
The Probit-LMM and its MAP approximation outperform both methods, because they contain both a dense kernel as well as a sparse linear effect.
Interestingly, even though the LMM-Lasso also has a sparse effect and a dense kernel, its performance is
not very compelling on our experimental dataset. This may be explained by its output being continuous
(and not binary), and therefore not well suited for classification tasks.

We also compared the runtimes across different methods, shown in Fig.~\ref{fig:time_bio_toy}. The Probit-LMM and
Probit regression have an approximately constant runtime in all scenarios whereas the latter is around $2.5$ times faster.
As expected, the runtime of Probit-LMM MAP lies between the other methods and slightly decreases in the more dense scenarios.
It can be considered a cheap alternative to the Probit-LMM, but predicts slightly worse.

Finally, we analyzed the importance of the $\ell_1$-norm regularizer in the Probit-LMM and compared it against a model that is $\ell_2$-regularized. We generated an artificial data set with $k=10$ non-zero features and tried to recover these non-zero feature weights with both algorithms. 
Fig.~\ref{fig:toy_sparsity} shows the results of this analysis. The blue solid line represents the truly non-zero weights, 
while the green dots show our estimates when using $\ell_1$-norm (left) and $\ell_2$-norm (right) regularization on $w$, respectively. We observe that the $\ell_1$-norm regularized Probit model finds better estimates of the linear weight vectors that were used to generate the data.

\subsection{Tuberculosis Disease Outcome Prediction}\label{sec:tub}

\begin{figure}[h!]
\includegraphics[width=0.49\linewidth]{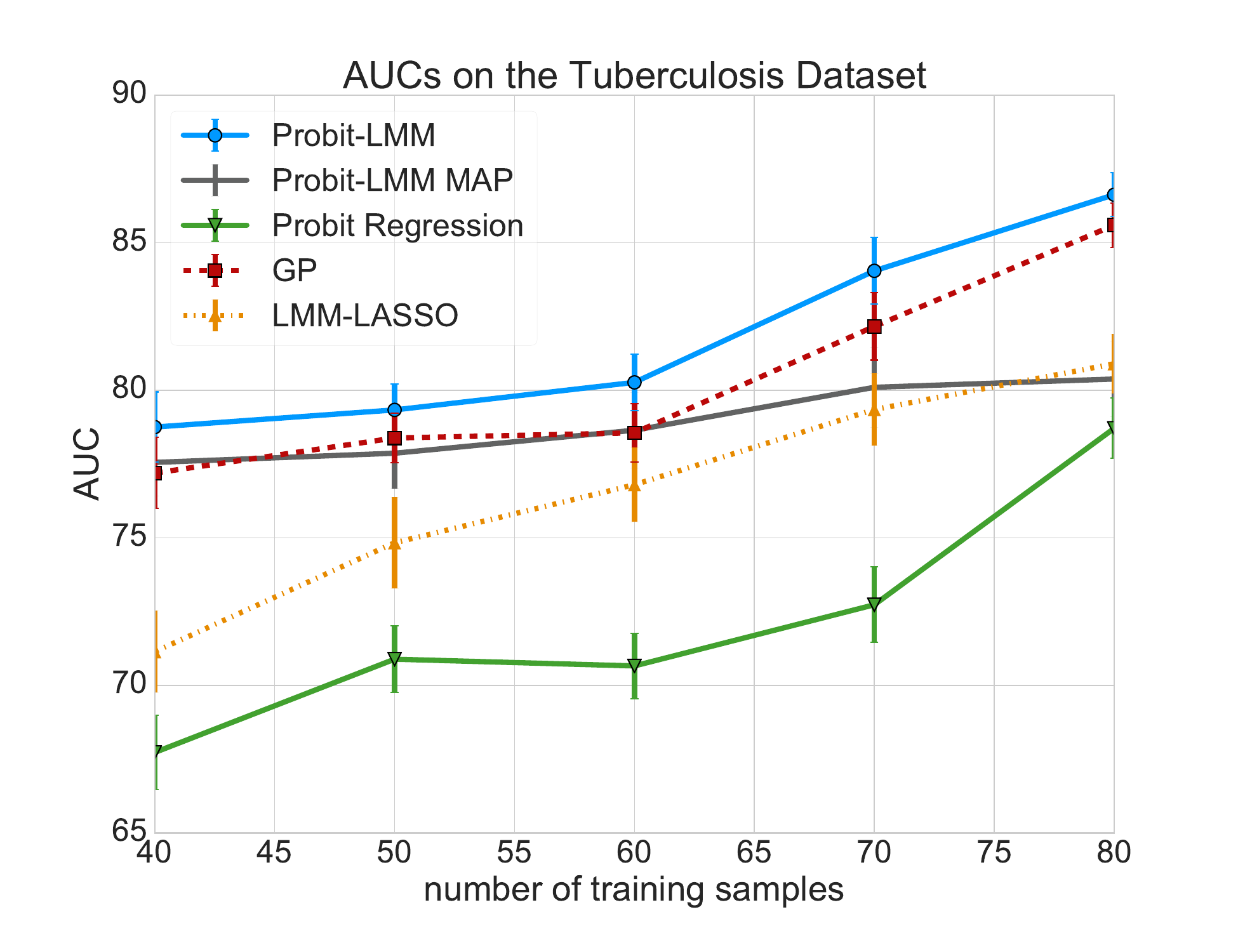} 
\vspace{0.2cm}
\includegraphics[width=0.49\linewidth]{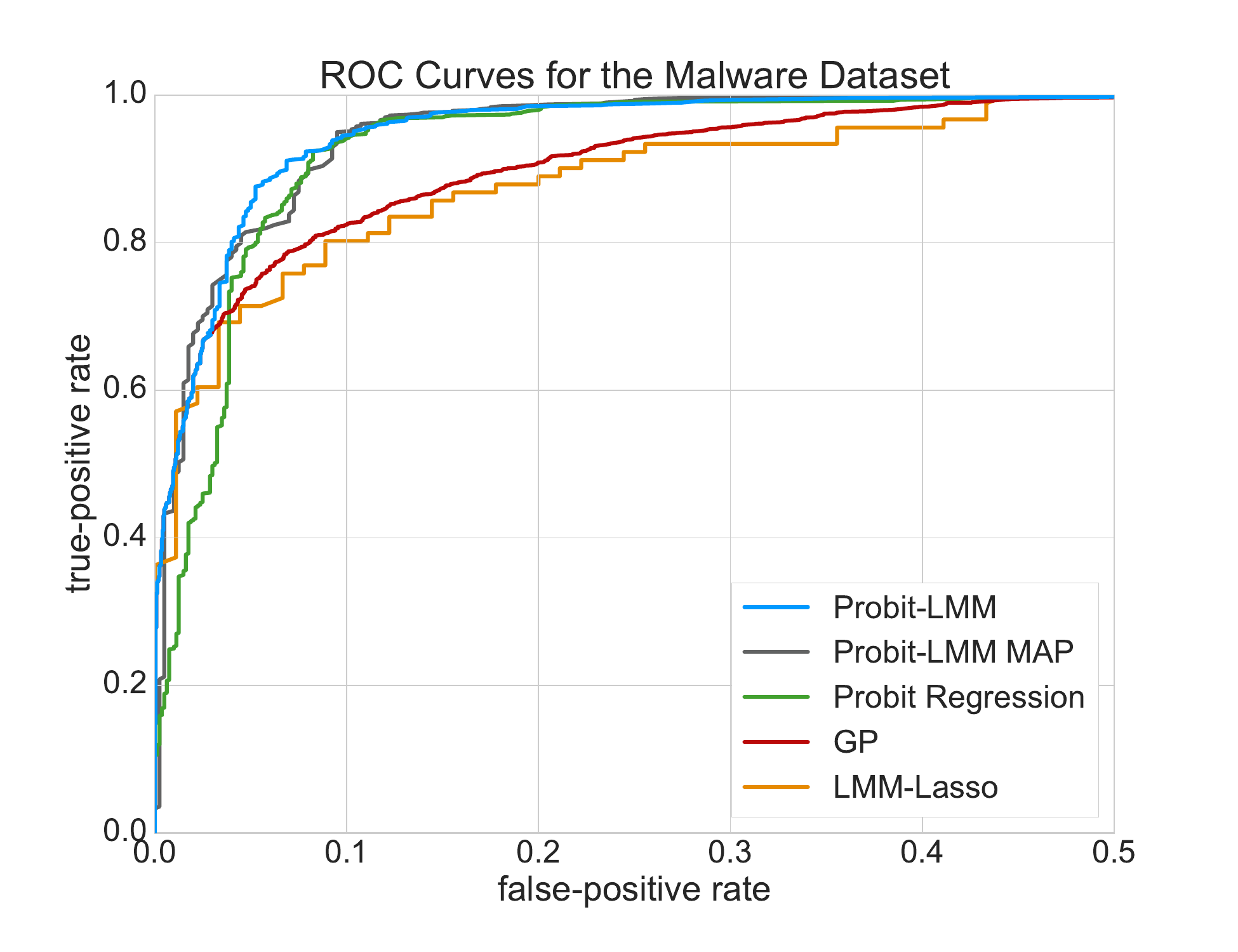} 
\caption{\textsc{Left}: Average AUC in the tuberculosis (TBC) experiment with respect to the training set size. 
  \textsc{Right}: Average ROC curves for the computer malware detection experiment. 
  \label{fig:auc_roc}}
\end{figure}

In our first real-world experiment, we predicted the outcome of Tuberculosis from gene expression levels.
We obtained the dataset by \citep{berry2010interferon} from the National Center for Biotechnology Information website\footnote{ {\small\url{http://www.ncbi.nlm.nih.gov/geo/query/acc.cgi?acc=GSE19491}}}, 
which includes $40$ blood samples from patients with active tuberculosis as well as $103$ healthy controls,
together with the transcriptional signature of blood samples measured in a microarray experiment with 48,803 gene expression levels, which
serve as features for our purposes. 
Also available is the age of the subjects when the blood sample was taken, from which we compute $\Sigma_{\text{side}}$\footnote{We compute $\Sigma_{\text{side}}$ as RBF kernel on top of the side information age using length scale $\sigma=0.2$.}. 
All competing methods are trained by using various training set sizes $n\in[40,80]$. To be consistent with previous studies (e.\,g. \citep{LiRakBor11}), we report on the area under the ROC curve (AUC), rather than accuracy.
The results are shown in Fig.~\ref{fig:auc_roc}, left.

\begin{figure}[!h]
\begin{center}
\includegraphics[width=0.4565\linewidth]{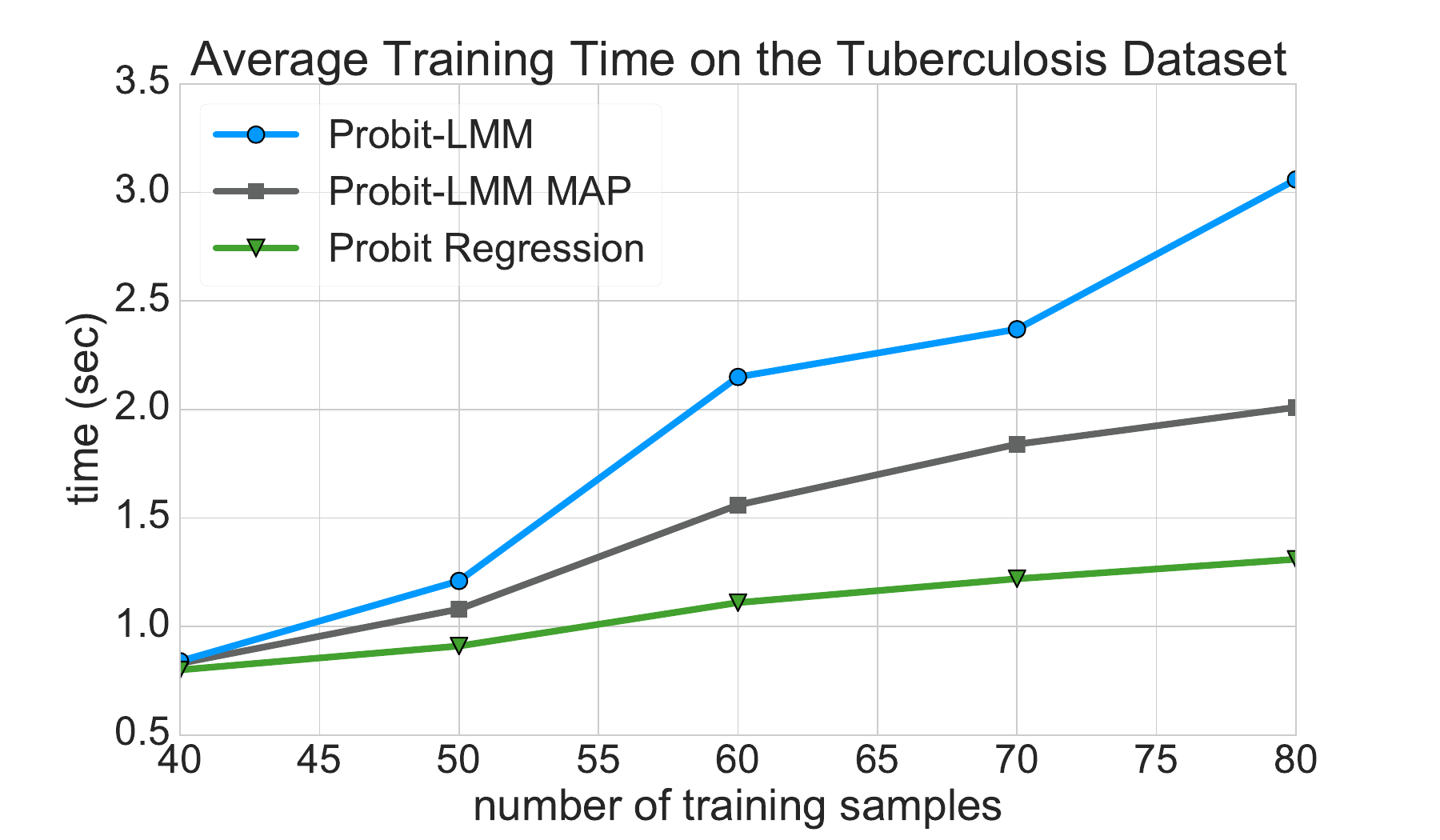}
\includegraphics[width=0.535\linewidth]{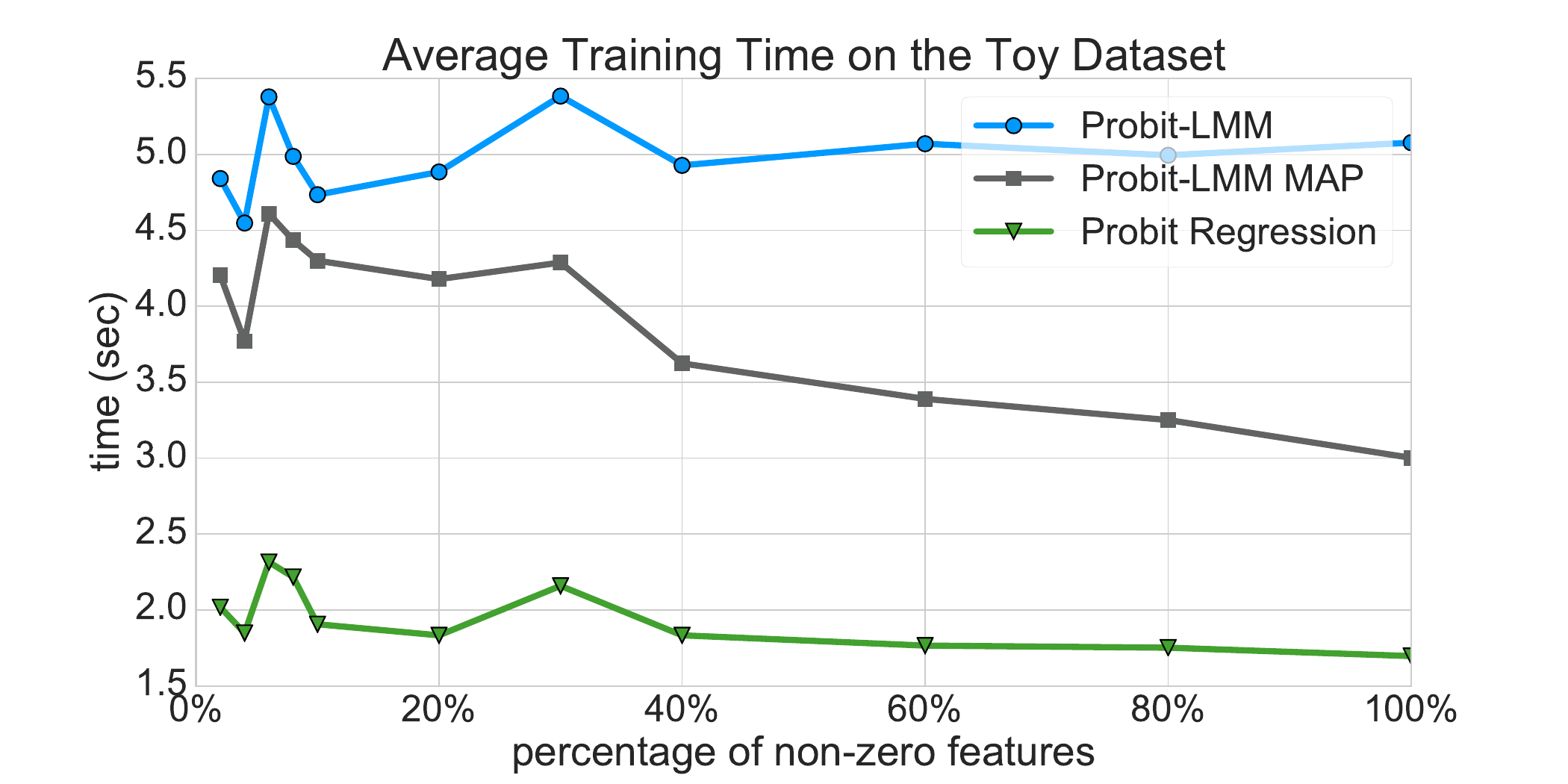}

\caption{
  \textsc{Toy}: Training time with respect to the dataset size in the tuberculosis experiment (\textsc{Left}) and with respect to the number of true non-zero features in the generating model (\textsc{Right}).
}
\label{fig:time_bio_toy}
\end{center}
\end{figure}

We observe that Probit-LMM achieves a consistent improvement over sparse Probit regression (by up to 12 percentage points),
GP classification (by up to $3$ percentage points), LMM-Lasso (by up to 7 percentage points) and its MAP approximation (by up to 7 percentage points).
In Fig.~\ref{fig:time_bio_toy} we show the runtime of Probit-LMM, its MAP version, and sparse Probit regression with respect to the dataset size. Note that both the prediction performance of the MAP approximation and its runtime lie between the full model (Probit-LMM) and sparse Probit regression.
In Fig.~\ref{fig:correlation}, left, we show the correlation between the top features and the population structure (as confounding factor) for the Probit-LMM and 
sparse Probit regression. The plot was created as explained in section \ref{sec:exp_setup}. We find that the features obtained by the Probit-LMM show less correlation with population structure than the features of sparse Probit regression. By inspecting the correlation coefficients of the first top 100 features of both methods, we observe that the features found by the Probit-LMM are less correlated with the confounder. This is because population structure was built into our model as a source of correlated noise.

\begin{figure}[ht]
\includegraphics[width=0.8\linewidth]{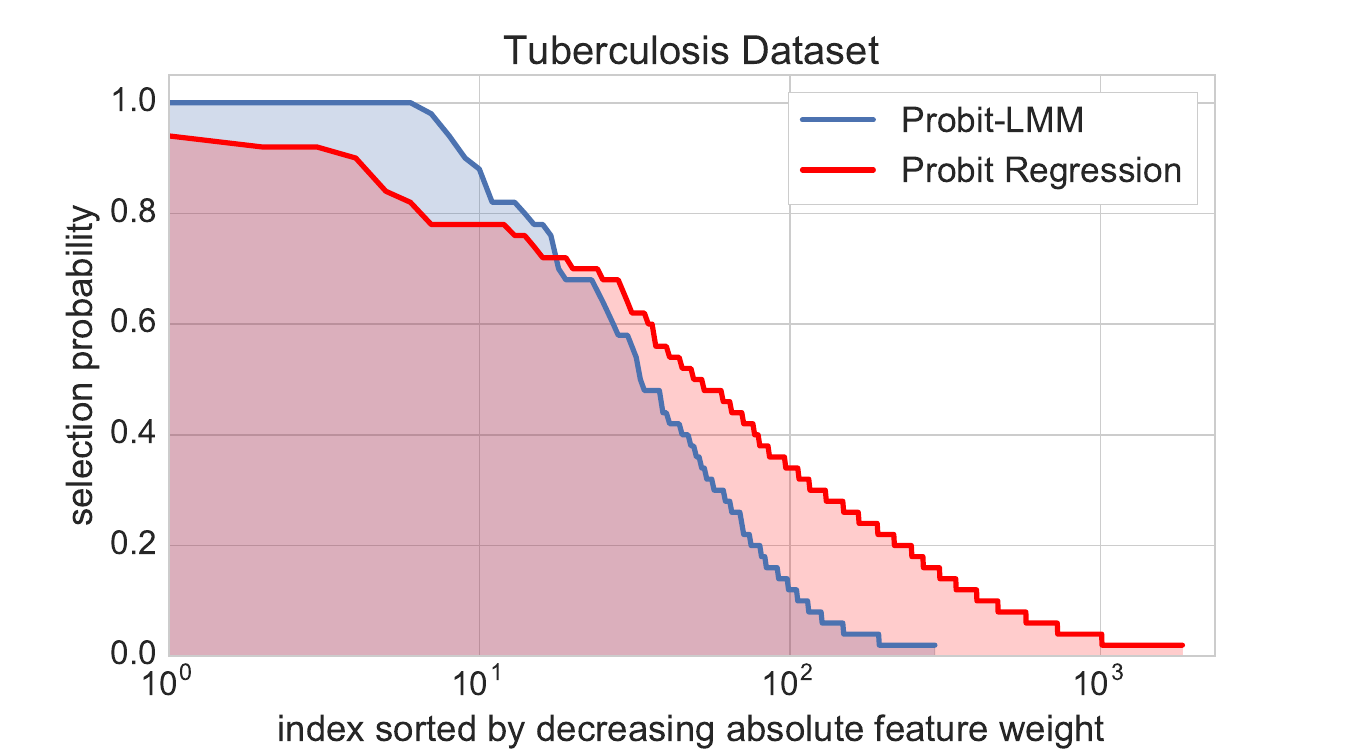} 
\caption{\textsc{TBC}: Stability of selected features for the Probit-LMM and sparse Probit regression. The plot shows the selection probabilities for each feature. Ideally, we want these to be $0$ or $1$. The Probit-LMM (proposed) leads to more stable top features and has less variability under bootstrapping.
  \label{fig:stability}}
\end{figure}

  To make sure that our selected features are reliable, we investigate their stability under bootstrapping.
  We considered stability selection \citep{for916}, where we randomly subsample $90\%$ of the data $100$ times
  (to accommodate the limited sample size, we follow \citep{Rakitsch13} and do not use $50\%$ of the samples for each draw as proposed in the original article).
  We define a feature to be selected if the absolute weight exceeds the threshold of $0.001$. 
  In Fig.~\ref{fig:stability} we show the selection probability for each feature. For the Probit-LMM, the top 7 features
  are selected in every singe run out of $100$ runs, indicating that they are very stable. In contrast, in standard
  sparse Probit regression (Lasso) these features only get selected with about $90\%$ probability. Also, the total number of
  selected features over all runs is 294 in our approach, whereas for sparse Probit regression it is 1837, which
  indicates that there is less variability compared to the standard Lasso approach. The Probit-LMM thus leads to more stable
  features than the standard Lasso approach since it also includes a dense effect as explained in section \ref{sec:pop}.

  Furthermore, we test the significance of the selected features of the Probit-LMM, where we construct a test statistic based
  on the likelihood ratio of our model and a reference model without fixed effect~\citep{neyman1933problem}. 
  Our null hypothesis is, thus, that these features do not influence the disease outcome, hence that a model where
  all these corresponding feature weights are zero is equally powerful.
  We train our method on $75\%$ of the data and valuate the likelihoods of both models on the remaining $25\%$ of the
  data and repeat this procedure $10$ times for random test-training splits. In each run, our algorithm selects between $32$ and $37$ features based on
  the aforementioned criterium that the feature weights exceed $0.001$. We obtain a log-likelihood ratio of $2.7 \pm 0.3$. Note that to construct a p-value out of this likelihood ratio, further assumptions about the distribution
  of model parameters would be required.

\subsection{Malicious Computer Software (Malware) Detection}\label{sec:malware}

We experiment on the Drebin dataset\footnote{\small\url{http://user.informatik.uni-goettingen.de/~darp/drebin/download.html}} \citep{arp2014drebin}, 
which contains 5,560 Android software applications from 179 different malware families.
There are 545,333 binary features; each feature denotes the presence or absence of a certain source code string
 (such as a permission, an API call or a network address).  
It makes sense to look for sparse representations~\citep{arp2014drebin}, as only a small number of strings are truly
characteristic of malware. The idea is that we consider populations of different families of malware when training, and hence correct for the analogue of genetic population structure in this new context,
that we call ``malware structure''.

\begin{figure}[h!]
\includegraphics[width=0.5\linewidth]{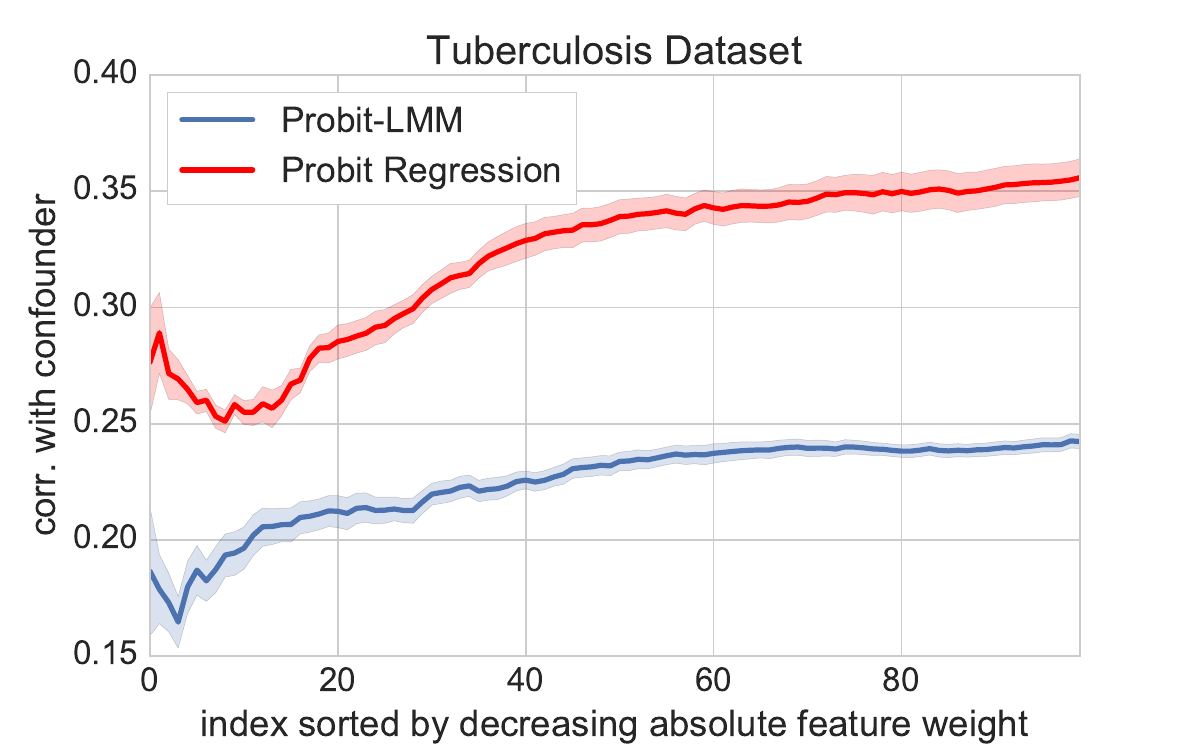} 
\vspace{0.2cm}
\includegraphics[width=0.5\linewidth]{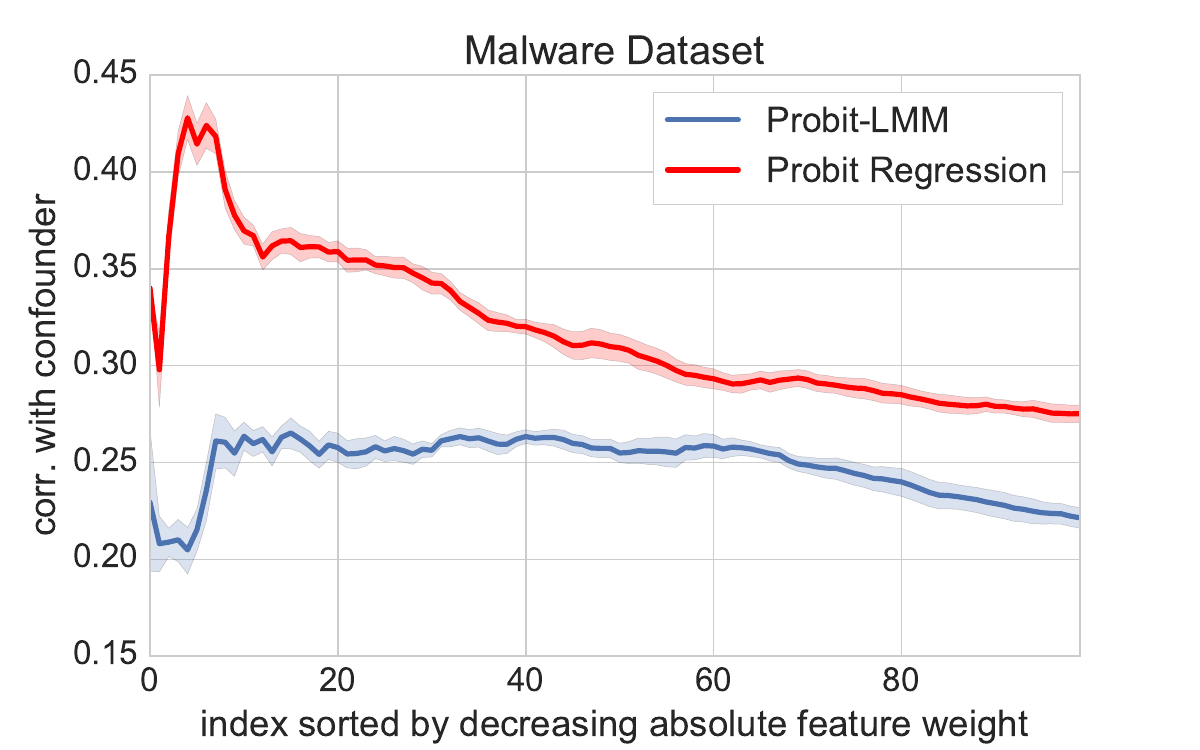} 
\caption{Correlation between the selected features and population structure as described in the main text (low values are better). The tuberculosis experiment is shown left, and computer malware shown right. The x-axis is sorted by descending absolute weights. Light-red/light-blue areas indicate standard errors.
 \label{fig:correlation} }
\end{figure}

We concentrate on the top 10 most frequently occurring malware families in the dataset.\footnote{Geinimi, FakeDoc, Kmin, Iconosys, BaseBridge, GinMaster, Opfake, Plankton, FakeInstaller, DroidKungFu.}.
We took 10 instances from each family, forming together a malicious set of 100 and a benign set of another 100 instances (i.e., in total 200 samples). 
We employ $n=80$ instances for training and stratify in the sense that we make sure that each training/validation/test set contains $50\%$ benign samples and an equal amount of malware instances from each family.
Since no side information is available, we only use a linear kernel and the identity matrix as components for the correlation matrix.
We report on the (normalized) area under the Receiver Operating Characteristic (ROC) curve over the interval $[0, 0.1]$ and denote this performance measure by $\text{AUC}_{0.1}$. In Fig.~\ref{fig:auc_roc}, right, we show the ROC curves, in Table~\ref{tab:drebin} the achieved  $\text{AUC}_{0.1}$
and in Table~\ref{tab:drebin_runtime} the runtimes of the Probit-LMM, its MAP approximation, and sparse Probit regression.

\begin{table}[h!]
\begin{center}
\begin{tabular}{|c|c|c|c|c|}
\hline
 Probit-LMM &
 Probit-LMM MAP & 
 Probit Regression&
 GP & 
 LMM-Lasso 
 \\ \hline
 $\mathbf{74.9\pm0.2}$&
 $73.1\pm0.4$  &
 $67.2\pm0.3$ & 
  $69.8\pm0.3$ &
   $66.45\pm0.3$  
   \\ \hline
\end{tabular}
\end{center}
\caption{\label{tab:drebin}\textsc{Malware:} $\text{AUC}_{0.1}$ and corresponding standard deviations attained on the malware dataset.}
\end{table}

\begin{table}[h!]
\begin{center}
\begin{tabular}{|c|c|c|}
\hline
 Probit-LMM &
 Probit-LMM MAP & 
 Probit Regression
 \\ \hline
 $14.89$ sec&
 $11.03$ sec&
 $8.91$ sec 
   \\ \hline
\end{tabular}
\end{center}
\caption{\label{tab:drebin_runtime} \textsc{Malware:} Average training time on the malware dataset.}
\end{table}

We observe that the Probit-LMM achieves a consistent improvement in terms of $\text{AUC}_{0.1}$ over sparse Probit regression 
(by approximately 7.5 percentage points),  GP classification (by approximately 5 percentage points), LMM-Lasso (by approximately 8.4 percentage points),
 and over its MAP approximation (by approximately 2 percentage points). Furthermore, in Fig.~\ref{fig:correlation}, right, we plot the correlation of the top features of Probit-LMM  and sparse Probit regression with population structure. We observe that the Probit-LMM leads to features which are less correlated with the malware structure.

\subsection{Flowering Time Prediction From Single Nucleotide Polymorphisms}\label{sec:flower}

We experiment on genotype and phenotype data consisting of 199 genetically different accessions (instances) from the model plant \emph{Arabidopsis thaliana} \citep{atwell2010genome}. 
The genotype of each accession comprises 216,130 single nucleotide polymorphism (SNP) features. 
The phenotype that we aim to predict is early or late flowering of a plant when grown at ten degrees centigrade. 
The original dataset contains the flowering time for each of the 199 genotypes. 
We split the dataset into the lower and upper $45\%$-quantiles of the flowering time and binarized the labels, resulting 
in a set of 180 accession from which we use $n=150$ accessions for training. 
The results are reported in Table~\ref{tab:flowering}
and show that the Probit-LMM has a slight advantage of at least $0.5$ percentage points in AUC over the competitors.
The MAP approximation can be considered as cheap alternative to the Probit-LMM since its prediction performance is only slightly worse than the Probit-LMM but it is substantially faster (see Table~\ref{tab:flower_runtime}).

\begin{table}[h!]
\begin{center}
\begin{tabular}{|c|c|c|c|c|}
\hline
 Probit-LMM & 
 Probit-LMM MAP &
 Probit Regression &
 GP &  
  LMM-Lasso 
    \\ \hline
 $\mathbf{84.1\pm0.2}$&
 $83.6\pm0.3$ &
 $83.5\pm0.2$ & 
  $83.6\pm0.2$ &
   $79.7\pm0.2$ 
   \\ \hline
\end{tabular}
\end{center}
\caption{\label{tab:flowering} \textsc{Flowering:} AUCs and corresponding standard errors in the flowering time prediction experiment.}
\end{table}

\begin{table}[h!]
\begin{center}
\begin{tabular}{|c|c|c|}
\hline
 Probit-LMM &
 Probit-LMM MAP & 
 Probit Regression
 \\ \hline
 $21.02$ sec&
 $13.17$ sec&
 $10.59$ sec 
   \\ \hline
\end{tabular}
\end{center}
\caption{\label{tab:flower_runtime} \textsc{Flowering:} Average training time in the flowering time experiment.}
\end{table}

An analysis restricted to the ten SNPs with largest absolute regression weights in our model showed that they lie within four well-annotated genes that all convincingly can be related to flowering, structure and growth: the gene AT2G21930 is a growth protein that is expressed during flowering, AT4G27360 is involved in microtubule motor activity, AT3G48320 is a membrane protein, involved in plant structure, and AT5G28040 is a DNA binding protein that is expressed during flowering.

\section{Conclusion}
\label{sec:conclusion}

We presented a novel algorithm for sparse feature
selection in binary classification where the training data show spurious correlations, e.g., due to confounding. 
Our approach generalizes the LMM modeling paradigm to binary classification, which poses technical challenges
as exact inference becomes intractable. Our solution relies on approximate Bayesian inference. 
We demonstrated our approach on a synthetic dataset and two data sets from the field of statistical genetics as well as
 third data set from the domain of compute malware detection.

Our approximate Bayesian EM-algorithm can be seen as a hybrid between an $\ell_1$-norm regularized Probit classifier (enforcing sparsity) and a GP classifier
that takes as input an arbitrary noise kernel.
It is able to disambiguate between sparse linear effects and correlated Gaussian noise and thereby explains away spurious correlations due to confounding. We showed empirically that our model selects features which show less correlation with the first principal components of the noise covariance, and which are therefore closer to the truly underlying sparsity pattern.

While sparsity by itself is not the ultimate virtue to be striven for, we showed that
the combination of sparsity-inducing regularization and dense-type probabilistic modeling (as in the proposed method) 
may improve over purely sparse models such as $\ell_1$-norm regularized (sparse) Probit regression.
The corresponding theoretical exploration is left for future work.
We note that a good starting point to this end will be to study the existing literature on compressed sensing as pioneered by \citep{Candes06,Donoho06} and put
forward by \citep{1bit-compressive} in the context of 1-bit compressed sensing. 
For the latter case such theory recently has been developed by \citep{Plan12}, but under the assumption of independent noise variables---an assumption that is violated in the Probit-LMM.

A shortcoming of the model is the fact that the noise covariance kernel is fixed in advance and is not
  learned from the data.
  As a possible extension, one could treat the design matrix $X$ which is used to compute the similarity kernel $K(X,X)$ as a free parameter and  optimize it according to a maximum likelihood criterion. For a linear kernel this would basically yield a probabilistic PCA, for a non-linear kernel such as in deep Gaussian processes or Gaussian process latent variable models, this can yield interesting forms of dimensionality reduction. However, these models are typically used to analyze higher dimensional data  where multiple outputs (phenotypes)
  per training example are available. Trying to estimate a covariance of size $n\times n$ with only $n$ training examples, we would run the danger of overfitting.
  This is also the reason why linear kernels of the feature matrix are still standard in genetics and are used in most LMM applications.

In the future, several paths are viable.
An interesting extension of our approach would be a fully Bayesian one that also captures parameter uncertainty over $w$.
To obtain the posterior on w, it might be easier to use sparsity-inducing hierarchical priors, e.g., an automatic relevance
determination prior or Gaussian scale mixture, instead of the Laplace prior. Second, multi-class versions of the model are possible.
And third, even more scalable approaches could be explored. To this end, one can make use of the formulation of the model in
Eq.~\ref{eq:global_local} and employ Stochastic Variational Inference, a scalable Bayesian algorithm based on stochastic
optimization~\citep{hoffman2013stochastic}.
We will leave these aspects for future studies.

\bibliographystyle{apalike}
\bibliography{refs}

\newpage
\appendix

\section{Absorbing the Label Signs by Preprocessing $X$ and $\Sigma$}

We have claimed in section~\ref{sec:2} that it is not a constraint to assume that all labels are $1$.
Hence, we show that the model $Y = {\rm sign}(X^\top w + \epsilon), \; \epsilon\sim {\cal N}(0,\Sigma)$ where $Y\equiv 1$
is indeed equivalent to another model $\tilde{Y} = {\rm sign}(\tilde{X}^\top w + \tilde{\epsilon}), \; \tilde{\epsilon}\sim {\cal N}(0,\tilde{\Sigma})$
where $\tilde{Y}$ is arbitrary.  We explicitly give the transformations between these two models and the corresponding variables. 

We start with the original problem where $\tilde{Y}  \in \{\pm 1\}^n$ is an arbitrary vector of binary labels.
The model furthermore involves the data matrix $\tilde{X} = (\tilde{X}_1,\cdots,\tilde{X}_n) \in {\mathbb R}^{d\times n}$
and a noise covariance $\tilde{\Sigma}$ such that
$\tilde{Y} = {\rm sign}(\tilde{X}^\top w + \tilde{\epsilon}), \; \tilde{\epsilon}\sim {\cal N}(0,\tilde{\Sigma}).$
We now transform every column of $\tilde{X}$ as $X_i = \tilde{X}_i \circ \tilde{Y}_i $, where $\circ$ is the Hadamard product. 
When multiplying this equation element-wise with $\tilde{Y}$, this yields
 $1 = \tilde{Y}\circ \tilde{Y}  = {\rm sign}(X^\top w + \tilde{Y}\circ \tilde{\epsilon}), \quad \tilde{\epsilon}\sim {\cal N}(0,\tilde{\Sigma}).$
Lastly, we observe that the random variable $\tilde{Y} \circ \tilde{\epsilon}$ with $\tilde{\epsilon}\sim {\cal N}(0,\tilde{\Sigma})$ has the same distribution
as $\epsilon$ with $\epsilon \sim {\cal N}(0, \Sigma )$ where we defined $\Sigma \equiv {\rm diag}(\tilde{Y}) \cdot \tilde{\Sigma} \cdot {\rm diag}(\tilde{Y}) $. 
To summarize,  after the above transformations,  the model reads
$
1 = {\rm sign}(X^\top w + \epsilon), \quad \epsilon \sim {\cal N}(0,\Sigma).
$
We see that we have effectively absorbed the arbitrary observed label $\tilde{Y}$ by means of a rotation of the data matrix and the noise covariance. This proves our claim.

\section{Convexity of the Objective Functions}
\label{sec:convexity}
We prove that the objective function Eq.~\ref{eq:CorrelatedProbit} and its MAP approximation Eq.~\ref{eq:MAP}  are convex.

We begin by proving convexity of Eq.~\ref{eq:CorrelatedProbit}.
Since the $\ell_1$-norm regularizer is convex it is sufficient to show that 
$
  {\cal L}_0(w) \equiv   -\log \int_{{\mathbb R}_+^n}  {\cal N}(\epsilon; X^\top w,\Sigma)\,d^n\epsilon \label{eq:convex_proof}
$
is convex in $w$. Recall that a function $f$ is  log-convex, if $f$ is strictly positive and $\log f$ is convex;  log-concavity is defined analogously. 
In the following, we make use of a theorem that connects log-concave functions to their partial integrals over convex sets~ \citep{Prekopa73}.
Namely, for a log-concave function $f:~{\mathbb R}^{n+m} \to {\mathbb R}$ and a convex subset $A \subset {\mathbb R}^n$, the function $g(x)=\int_{A}f(x,y) d^m y$ is log-concave in the entire space ${\mathbb R}^n$.
Since $X^\top w$ is linear, it is sufficient  to show that
$
  f(\mu) := -\log \int_{{\mathbb R}_+^n}  {\cal N}(\epsilon; \mu,\Sigma) \, d^n \epsilon
$
is convex in $\mu$. The multivariate Gaussian density ${\cal N}$ is log-concave in $(\epsilon, \mu) \in {\mathbb R}^{2n}$, since ${\cal N}(\epsilon; \mu,\Sigma)>0$ for all $\mu, \epsilon \in {\mathbb R}^n$ and $\log {\cal N}$ is concave in $(\epsilon, \mu)$. Therefore, $\int_{{\mathbb R}_+^n}  {\cal N}(\epsilon; \mu,\Sigma) \, d^n \epsilon$ is log-concave in $\mu$. The logarithm of a log-concave function is concave by definition. Thus,
$f$ is convex in $\mu$ and therefore, Eq.~\ref{eq:CorrelatedProbit} is convex in $w$~\qed.

Let us now consider the objective function of the MAP approximation, Eq.~\ref{eq:MAP}. Since the regularizers are convex in $w$ and $w'$, it is sufficient to show that
$
 - \sum_{i=1}^n \log \Phi\left(X_i^\top(w+w') / \sqrt{\lambda_1}\right) \label{eq:convexity_proof2}
$
is convex in $(w,w') \in {\mathbb R}^{2n}$. With analogous arguments showing the convexity of $f(\mu)$, it holds that 
$
  g(\mu):= \log \Phi(\frac{\mu}{\sqrt{\lambda_1}})
$
is convex in $\mu$. Since $X_i^\top (w +w')$ is linear in $(w,w')$, it follows that Eq.~\ref{eq:MAP} is convex in $(w,w')$~\qed.

\section{Predicting New Labels}
\label{sec:prediction}
When predicting new labels in the Probit-LMM, we have two choices.
We can either ignore correlations between samples, or take them into account.
Both cases have their use which depends on the context. While in the first case we
simply take the sign of $X^\top w$ of a new data point to predict its label, the second
case closely resembles prediction in Gaussian Processes~\citep{book:Rasmussen+Williams:2006}
and shall here be reviewed.

We introduce letters that indicate the t{\bf r}aining set (R) and the t{\bf e}st set (E),
and let $y_{E/R}$ be the test and training labels, respectively. 
We define the mapping
$
Y_E \mapsto Y := (Y_E^\top ,Y_R^\top )^\top \in {\mathbb R}^{m+n}. 
$
We also concatenate test data and training data as
$
X = (  X_E^\top , X_R^\top )^\top \in {\mathbb R}^{d \times (m+n)}.
$
Finally, we consider the concatenated kernel matrices
\be
K^i = \begin{pmatrix}  K^i_{EE} & K^i_{ER} \\  K^i_{RE} &  K^i_{RR} \end{pmatrix} \in {\mathbb R}^{(m+n) \times (m+n)}
\ee
We use the weights $\lambda_i$ that were determined by model selection on the training data $(Y_R,X_R)$ to construct the covariance matrix on the extended space,
$
\Sigma = \sum_i \lambda_i K^i.
$
In order to predict new labels $Y_E$, we evaluate the objective, using $X$, $Y = Y(Y_E)$ and the training weights $w$.
The predicted label is then 
$
Y_E^{*}  =  {\rm arg} \min_{Y_E \in \{\pm 1\}^m } {\cal L}(w| X, Y,\Sigma). \nonumber
$

\section{Gradient and Hessian}\label{grad_hess}
In this section, we calculate the gradient and the Hessian of the un-regularized objective,
$
{\cal L}_0(w)  =  - \log \int_{{\mathbb R}_+^n} {\cal N}(\epsilon; \mu(w),\Sigma )d^n \epsilon.
$
It will be sometimes more convenient  to consider the objective as a function of $\mu=X^\top w$, rather than $w$,
for which case we define
$
{\cal L}_0(\mu)  =  - \log \int_{{\mathbb R}_+^n} {\cal N}(\epsilon; \mu,\Sigma )d^n \epsilon.
$
We begin by computing the gradient. We define $\mu_p =  {\mathbb E}_{p(\epsilon|\mu,\Sigma)}\left[\epsilon \right] $ as the mean of  the truncated Gaussian.
The gradient is given by
$$
\nabla_{w} {\cal L}_0(w)   =   \frac{ \int_{{\mathbb R}_{+}^n } (\epsilon - \mu)^\top \Sigma^{-1}\,{\cal N}(\epsilon; \mu, \Sigma) d^n\epsilon}{ \int_{{\mathbb R}_+^n} {\cal N}(\epsilon; \mu, \Sigma)d^n\epsilon } \,X^\top  =  (\mu_p - \mu)^\top \Sigma^{-1} X^\top. 
$$
We now compute the Hessian. 
We first consider  the Hessian matrix of ${\cal L}_0(\mu)$,
$
B_{ij}(\mu)  =   \partial_{\mu_i}\partial_{\mu_j}{\cal L}_0(\mu).
$
The chain rule relates this object to the Hessian of ${\cal L}_0(w)$, namely
$
H(w) =  X  B(\mu) X^\top.
$
The problem therefore reduces to calculating $B(\mu)$ which is  $n\times n$,
whereas the original Hessian $H(w)$ is $d \times d$. 

To calculate $B(\mu)$, we define
$I(\mu) =  \int_{{\mathbb R}^+_n} \exp \{ -\frac{1}{2} (\epsilon-\mu)^\top \Sigma^{-1} (\epsilon-\mu) \} d^n\epsilon$. 
Up to a constant, ${\cal L}_0(\mu) =- \log I(\mu)$.
The Hessian is given by
$
B_{ij}(\mu)  =  -\frac{\partial_{\mu_i}  \partial_{\mu_j} I(\mu)}{ I(\mu)} +  \frac{\partial_{\mu_i}  I(\mu)}{ I(\mu)} \frac{ \partial_{\mu_j} I(\mu)}{ I(\mu)}.
$
Note that this involves also the first derivatives of $I(\mu)$, that we have already calculated for the gradient. 
To proceed, we still need to calculate $\partial_{\mu_i}  \partial_{\mu_j} I(\mu)$. To simplify the calculation, we introduce
$
\tilde{\mu}  =  \epsilon - \mu.
$
As a consequence, $\partial_{\tilde{\mu}_i} = -\partial_{\mu_i}$. Furthermore,
$$
\partial_{\mu_i}  \partial_{\mu_j} \exp\{ -\frac{1}{2}(\epsilon - \mu)^\top \Sigma^{-1}(\epsilon-\mu)  \} =  \left[\Sigma^{-1} \tilde{\mu} \tilde{\mu}^\top\Sigma^{-1} - \Sigma^{-1}\right]_{ij} \exp\{ -\frac{1}{2}\tilde{\mu}^\top \Sigma^{-1}\tilde{\mu}  \} .
$$
Based on this identity, we derive
$
\frac{\partial_{\mu_i}  \partial_{\mu_j}  I(\mu)}{I(\mu)} =  \left( \Sigma^{-1} \Sigma_p\Sigma^{-1}   - \Sigma^{-1}  \right)_{ij}. 
$
For the remaining terms, we  use our known result for the gradient, namely
\be
 \frac{\partial_{\mu}  I(\mu)}{ I(\mu)}  =  \left( {\mathbb E}_{p(\epsilon|\mu)} [(\mu_p - \mu)^\top \Sigma^{-1} ] \right) = (\mu_p - \mu)^\top \Sigma^{-1}\nonumber.
\ee
As a consequence, 
\be
 \frac{\partial_{\mu_i}  I(\mu)}{ I(\mu)}   \frac{\partial_{\mu_j}  I(\mu)}{ I(\mu)}  =  \left(\Sigma^{-1}\Delta \mu\, \Delta \mu^\top  \Sigma^{-1}  \right)_{ij} \nonumber.
\ee
Above we defined $\Delta \mu = (\mu - \mu_q)$. This lets us summarize the Hessian matrix  $B(\mu)$:
\be
B(\mu) & = &  \left[ \Sigma^{-1} (\Sigma_p - \Delta \mu\, \Delta \mu^\top) \Sigma^{-1}   - \Sigma^{-1}  \right] \n
\ee
This gives us the Hessian.

\paragraph{Hessian Inversion Formula.}

For the second order gradient descent scheme, we need to compute the inverse matrix of the Hessian $H(w)$.
Let us call $D = \lambda_0 {\mathbf I}_{n}$ the (diagonal) Hessian of the regularizer.
We use the Woodbury matrix identity,
\be
H^{-1} & = & (D + XB X^\top)^{-1} \\
     & = & D^{-1} - D^{-1} X(B^{-1} + X^\top D^{-1} X)^{-1}  X^\top D^{-1} \n
             & = & \lambda_0^{-1}{\mathbf I}_{n} \lambda_0^{-2}  X(B^{-1} + \lambda_0^{-1} X^\top X)^{-1}  X^\top.  \nonumber
\ee
Note that this identity does not require us to invert a $d\times d$ matrix, but only involves the inversion of $n\times n$ matrices
(in our genetic applications, the number of samples $n$ is typically in the hundreds, while the number of genetic features $d$ in 
is of order $10^4-10^5$).
We first precompute the
linear kernel $X^\top X$. We also use the fact that we can more efficiently compute the product $H^{-1} \nabla_w {\cal L}$ as opposed to first
calculating the Hessian inverse and then multiplying it with the gradient.

\end{document}